\def\year{2022}\relax

\documentclass[letterpaper]{article} 
\usepackage{aaai22}  
\usepackage{times}  
\usepackage{helvet}  
\usepackage{courier}  
\usepackage[hyphens]{url}  
\usepackage{graphicx} 
\usepackage{natbib}  
\usepackage{caption} 
\DeclareCaptionStyle{ruled}{labelfont=normalfont,labelsep=colon,strut=off} 
\usepackage{todonotes}
\usepackage{subfigure}

\usepackage{latexsym}
\usepackage{amssymb, amsmath}
\usepackage{tabularx}
\usepackage{booktabs}

\usepackage{newfloat}
\usepackage{listings}
\usepackage[inline]{enumitem}   
\makeatletter
\newcommand{\inlineitem}[1][]{%
\ifnum\enit@type=\tw@
    {\descriptionlabel{#1}}
  \hspace{\labelsep}%
\else
  \ifnum\enit@type=\z@
   \refstepcounter{\@listctr}\fi
    \quad\@itemlabel\hspace{\labelsep}%
\fi}

\makeatletter
\newif\if@restonecol
\makeatother

\usepackage[linesnumbered,ruled,vlined]{algorithm2e}
\usepackage{algpseudocode}

\mathchardef\mhyphen="2D

\newcommand{\Attention}{\mathrm{A{\scriptstyle TTENTION}}}
\newcommand{\Combined}{\mathsf{Combined}}
\newcommand{\Concat}{\oplus}
\newcommand{\DeepQ}{\mathsf{DeepQR}}
\newcommand{\EDFsolo}{\mathsf{EDF\mhyphen Solo}}
\newcommand{\EDFenriched}{\mathsf{EDF\mhyphen Enriched}}
\newcommand{\embed}{\mathrm{em}}
\newcommand{\FeedForward}{\mathrm{F{\scriptstyle EED}F{\scriptstyle ORWARD}}}
\newcommand{\head}{\mathrm{head}}
\newcommand{\MultiHead}{\mathrm{M{\scriptstyle ULTI}H{\scriptstyle EAD}}}
\newcommand{\linear}{\mathrm{lin}}

\newcommand{\R}{\mathbb{R}}
\newcommand{\RoBERTa}{\mathsf{RoBERTa}}
\newcommand{\rd}{\mathrm{rd}}
\newcommand{\re}{\mathrm{Re}}
\newcommand{\SBERT}{\mathsf{SBERT}}
\newcommand{\SCQC}{\mathsf{SCQC}}
\newcommand{\selfatt}{\mathrm{S{\scriptstyle ELF}A{\scriptstyle TTENTION}}}
\newcommand{\SFglove}{\mathsf{SF}}
\newcommand{\softmax}{\mathrm{softmax}}
\newcommand{\trans}{\mathrm{tr}}
\newcommand{\transpose}{\mathrm{T}}

\newtheorem{definition}{Definition}

\title{$\DeepQ$: Neural-based Quality Ratings for Learnersourced Multiple-Choice Questions}

\author{Lin Ni\textsuperscript{1}, Qiming Bao, Xiaoxuan Li, Qianqian Qi, \\
Paul Denny, Jim Warren, Michael Witbrock, Jiamou Liu \textsuperscript{2} \\
} 
\affiliations{
    School of Computer Science, The University of Auckland, New Zealand \\
    \{l.ni\textsuperscript{1}, jiamou.liu\textsuperscript{2}\}@auckland.ac.nz
}

\begin{document}

\maketitle

\begin{abstract}
Automated question quality rating (AQQR) aims to evaluate question quality through computational means, thereby addressing emerging challenges in online learnersourced question repositories. 
Existing methods for AQQR rely solely on explicitly-defined criteria such as readability and word count, while not fully utilising the power of state-of-the-art deep-learning techniques. 
We propose $\DeepQ$, a novel neural-network model for AQQR that is trained using multiple-choice-question (MCQ) datasets collected from PeerWise, a widely-used learnersourcing platform. Along with designing $\DeepQ$, we investigate models based on explicitly-defined features, or semantic features, or both. We also introduce a self-attention mechanism to capture semantic correlations between MCQ components, and a contrastive-learning approach to acquire question representations using quality ratings. Extensive experiments on datasets collected from eight university-level courses illustrate that $\DeepQ$ has superior performance over six comparative models.
\end{abstract}

\section{Introduction}
%
Recent shifts towards online learning at scale have presented new challenges to educators, including the need to develop large repositories of content suitable for personalised learning and to find novel ways of deeply engaging students with such material \cite{dhawan2020online,davis2018activating}.  Learnersourcing has recently emerged as a promising technique for addressing both of these challenges \cite{kim2015learnersourcing}.  Akin to crowdsourcing, learnersourcing involves students in the generation of educational resources. In theory, students benefit from the deep engagement needed to generate relevant learning artefacts which leads to improved understanding and robust recall of information, a phenomenon known as the generation effect \cite{crutcher1989cognitive}. In addition, the large quantity of resources created from learnersourced activities can be used by students to support regular practice, which
is known to be a highly effective learning strategy \cite{roediger2006test,carrier1992influence}, especially when spaced over time and when feedback is provided \cite{kang2016policy}.


Despite the well-established benefits of learnersourcing 
for students \cite{moseley2016impact,ebersbach2020comparing}, evaluating and maintaining the quality of student-generated repositories is a significant challenge \cite{walsh2018formative,moore2021examining}.  
Low quality content can dilute the value of a learnersourced repository, and negatively affect students' perceptions of its usefulness.  On the other hand, the identification of high-quality content can facilitate useful recommendations to students when practicing.  
Therefore, approaches for accurately assessing the quality of student-generated resources are of great interest.  Involving domain experts in the evaluation process is very costly and does not scale, negating one of the key benefits of learnersourcing.  A more scalable solution is to have students review and evaluate the content themselves \cite{darvishi2021employing}.  Prior research has shown that students can make similar quality judgments to experts, especially when the assessments provided by multiple students are aggregated \cite{abdi2021evaluating}.  
However, a sufficient number of students must view and evaluate each artefact before a valid assessment can be produced, which is inefficient. Computing 
quality assessments of content, at the moment it is produced, would benefit all learners.
Such {\em a priori} assessment of quality remains a difficult yet important challenge in learnersourcing contexts.



Current learnersourcing tools support a wide variety of artefact types, including hints, subgoal-labels, programming problems and complex assignments \cite{piotr2015learnersourcing, kim2013learnersourcing, leinonen2020crowdsourcing, pirttinen2018crowdsourcing, denny2011codewrite}. Multiple-choice questions (MCQs) are a very popular format in learnersourcing platforms, appearing in tools such as RiPPLE \cite{khosravi2019ripple}, Quizzical \cite{riggs2020positive}, UpGrade \cite{wang2019upgrade} and PeerWise \cite{denny2008peerwise}. Hence, a generalisable model that can assess the quality of student-generated MCQs has the potential for a significant impact.

In this work, we explore the problem of {\em automated question quality rating} (AQQR) by developing a computational method to rate student-generated MCQs {\em a priori}.
Existing measures of MCQ quality target explicitly-defined features such as cognitive complexity with respect to Bloom's taxonomy \cite{bates2014assessing},  
the justification of rationales \cite{choi2005scaffolding}, 
and the feasibility of distractors  \cite{papinczak2012using,galloway2015doing}.  Such measures require costly and subjective manual evaluation by experts. Recent progress in natural language processing (NLP) provides a suite of tools for extracting and analysing rich features of texts, presenting a real opportunity to enhance existing measures of quality.

\begin{figure}[h]
    \centering
    \includegraphics[width=0.35\textwidth]{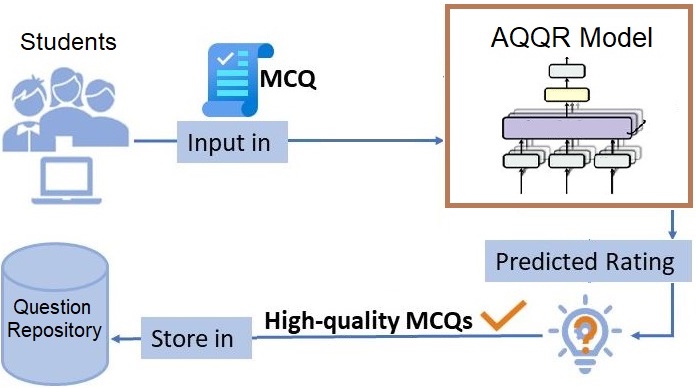}
    \caption{\footnotesize $\DeepQ$ used in a learnersourcing platform to remove low-quality student-generated questions.}\label{fig:Deep AQQR}
\end{figure}

\paragraph*{\bf Contribution.} We propose $\DeepQ$, a novel neural network-based model for AQQR that is trained using datasets collected from student-generated MCQ repositories. $\DeepQ$  is designed to be used in learnersourcing platforms to provide useful and immediate feedback to students and instructors.  Figure~\ref{fig:Deep AQQR} illustrates a potential use of $\DeepQ$ in practice. To the best of our knowledge, this is the first work that employs deep learning techniques to produce ratings of question quality. The design of $\DeepQ$ is guided by three research goals: (1) 
Capitalise on existing work that has identified indicative criteria for question quality -- such as readability or word count -- and utilise these for AQQR using tools from NLP.
In particular, we investigate the extraction of correlations between various {\em MCQ components}: the question stem, the correct answer, the distractors and explanation. (2)  Given the immense success of neural-based models for natural language understanding, it is natural to consider their application to the extraction of meaning from MCQs. The second research goal thus seeks to utilise rich semantic features to solve AQQR. (3) The research goals above suggest two sources of input features that are potentially useful for AQQR, namely, the explicitly-defined features (EDF) discussed in (1), and the semantic features (SF) discussed in (2). The third research goal is to explore their ``interplay'', i.e., how combining EDF with SF could facilitate a superior model for AQQR. 

To answer (1), we propose two complementary models: the first is an AQQR model that takes 18 EDFs as input including word counts, clarity, correctness, and readability indices. Since these features were not designed to capture relations between different question components, in the second model, we employ a self-attention mechanism that discovers {\em semantic-based correlations of question components} (SCQC). We demonstrate that enriching the input features in the first model with SCQC drastically improves AQQR performance.  To answer (2), we design a model that feeds GloVe embeddings into a transformer to produce a representation that captures the semantics of an MCQ. We demonstrate that this semantic feature-based model solves AQQR with better performance than benchmark models such as $\RoBERTa$ and $\SBERT$ which are much more costly to train, and at a comparable performance as the second model designed for (1). This demonstrates the value of semantic features (SF) in estimating question quality.  To answer (3), we propose the $\DeepQ$ model by combining EDF, SCQC, and SF as discussed above. Furthermore, to improve performance, we introduce a {\em quality-driven question embedding} (QDQE) scheme, which employs contrastive learning to fine-tune GloVe embeddings and better reflect question quality. This embedding is then used to enhance our model, providing a considerable performance boost. 
Our experiments were conducted using eight datasets (comprising 15,350 questions and more than 1,000,000 student-assigned quality ratings) from the PeerWise \cite{denny2008peerwise} learnersourcing platform, collected from medicine and commercial law courses. For most courses, our model is able to achieve accuracy in excess of 80\% (see Section~\ref{sec:experiment}). We also conduct a systematic analysis to validate our design decisions and the applicability of our model.



\section{Related Work}

Evaluating the quality of questions has attracted significant interest in educational research.  Standard metrics from classical test theory and item response theory, such as discrimination indices, have been used for decades to aid instructors in identifying poor quality questions \cite{brown2021setting, malau2014using}.  However, computing such quantitative measures requires large quantities of data on student responses to items.  Various qualitative indicators have also been proposed and used, such as {\em question clarity} \cite{choi2005scaffolding} and {\em distractor-plausibility} \cite{bates2014assessing}.  
Prior studies assessing such qualitative measures, however, have involved significant manual effort by experts.
In learnersourcing contexts, aggregating student ratings to assess the quality of questions is scalable and agrees well with expert ratings of quality \cite{abdi2021evaluating, darvishi2021employing,mcqueen2014peerwise}.  In our own work, we use averaged student ratings as ground-truth labels when training our AQQR models.

AQQR has recently attracted the attention of the machine learning community. In particular, the inaugural {\sc NeurIPS 2020 Education Challenge} included a quality prediction task for mathematics questions \cite{wang2021results}.  
Two of the successful entries in the NeurIPS challenge relied on {\em explicitly-defined features} (EDF), such as difficulty and readability, deriving a final rating using some form of average over these feature values \cite{shinahara_takehara_2020, tal_machine_learning_team_2020}.
Such approaches are limited in the sense that they rely on ad-hoc EDFs and linear transformations which may not provide the level of robustness and flexibility required for a diverse range of questions and courses.  
Neural networks are able to extract distributional semantics from texts offering greater richness and versatility.  Yet, to our knowledge, there has not been a systematic effort to design neural-based models to rate question quality.  Our paper aims to fill this gap by investigating the value of {\em semantic features} (SF) in AQQR. 
We note that the question dataset published for the {\sc NeurIPS 2020 Education Challenge} is {\em not} suitable for our purpose as the questions are largely mathematical and involve diagrams.

Tasks that closely resemble AQQR -- and for which neural network-based approaches have proven useful -- include {\em question difficulty prediction} (QDP) and {\em automated essay scoring} (AES).  QDP requires evaluation of a difficulty score for reading comprehension questions. \citet{huang2017question} approached this task using deep learning with an attention-based CNN model, and subsequent work by \citet{qiu2019question} incorporated a knowledge extraction aspect.  Both works rely heavily on the extraction of rich SFs from a question to predict its difficulty.  AES seeks to rate an article's quality based on its content, grammar, and organization.  Early AES models generally applied regression methods to a set of EDFs \cite{shermis2003automated}. \citet{aes2016} was the first to tackle AES using deep learning by automating feature extraction using a combination of convolutional and recurrent neural networks.  More recently, \citet{uto2020neural} combined EDF input and SF extracted from the pre-trained language model Bidirectional Encoder Representations from Transformers ($\mathsf{BERT}$).

Pre-trained language models have brought major breakthroughs with significant performance improvement and training cost savings in NLP \cite{bommasani2021opportunities}. Relying on its powerful language representation ability and easy scalability for various downstream tasks, $\mathsf{BERT}$\cite{DBLP:journals/corr/abs-1810-04805} and its extended models often appear at the forefront of the NLP benchmark leaderboards. Among them, $\RoBERTa$ \cite{liu2019roberta} improves $\mathsf{BERT}$ by pre-training on a larger dataset with more parameters; while Sentence-BERT($\SBERT$) \cite{reimers-2019-sentence-bert} is trained using siamese $\mathsf{BERT}$-Networks on paired sentences to derive better semantic embeddings. Both of them outperform $\mathsf{BERT}$ in well-established benchmark tasks.

\section{Problem Formulation}\label{sec:problem}
Here we formally define the {\em automated question quality rating} (AQQR) task. Each PeerWise dataset that specifies an instance of the task contains student-authored questions for a university course. When authoring an MCQ, the student specifies seven components: a question stem, a correct answer, (up to) four distractors, and a paragraph that explains the idea and rationale behind the question. 
The question is then submitted to an online question repository accessible by the class. After answering a question, a student may leave a holistic quality rating (from $0,1,\ldots,5$) by considering the  
``\textit{language, quality of options, quality of explanation, and relevance to the course}'' as suggested by the system.
We provide a sample MCQ below: 

\noindent\fbox{%
    \parbox{0.46\textwidth}{\scriptsize %
        \begin{itemize}[noitemsep,nolistsep,leftmargin=*]
        \item {\bf Stem:} Mr. Cram-zan is chilling in his room wondering another new way in which to make money. He believes he should create a global footballing league as God is telling him to. He is the chosen one, not Mourinho. He also thinks his close friend, Moo Leerihan, is plotting the downfall of his league. What is Mr. Cram-zan suffering from?
        \item {\bf Answer:} Schizophrenia
        \item {\bf Distractor 1:} Hallucinations \inlineitem {\bf Distractor 2:} Illusions
        \item {\bf Distractor 3:} Over ambition \inlineitem {\bf Distractor 4:} Being too chilled
        \item {\bf Explanation:} Schizophrenia would be the SBA as it encompasses all the aspects.
        \item {\bf Average rating:} 2.71
        \end{itemize}
    }%
}

\begin{definition}(AQQR) Given a set of MCQ $M_1,M_2,\ldots,M_n$ collected from a course, where each $M_i$ consists of a stem $S_i$, a correct answer $A_i$, distractors $D_{i,j}$ where $j\in \{1,2,3,4\}$, explanation $E_i$, and is assigned a rating $r_i$, {\em AQQR} seeks to build a prediction model $\mathsf{Rate}$ that estimates the rating of MCQ in the newly-conducted test set. 
\end{definition}

\begin{figure}
    \centering
    \includegraphics[width=0.4\textwidth]{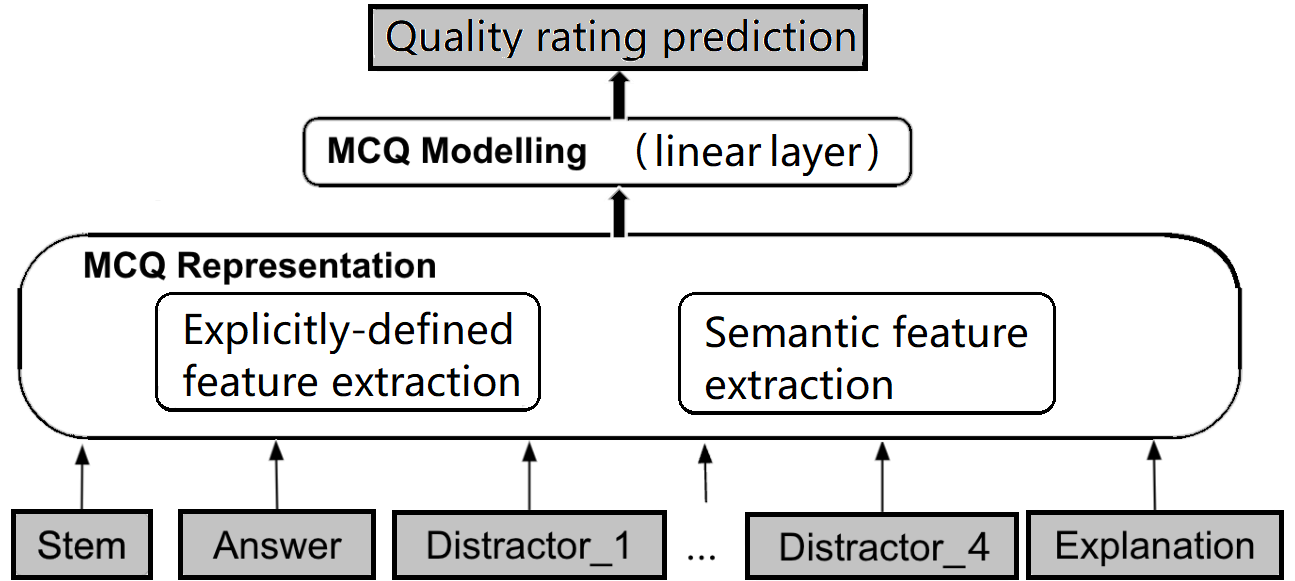}
    \caption{\footnotesize An overview of AQQR subtasks. }\label{fig:model}
\end{figure}

We view AQQR as consisting of two subtasks: (1) {\em MCQ representation} aims to 
process the input data to form feature vectors. The features are manually specified or automatically extracted. The former extracts {\em explicitly-defined features} (EDF) by leveraging domain knowledge and expert judgement while the latter captures {\em semantic features} (SF) using machine learning algorithms. We speculate that both types of features could be useful for our task. (2) {\em MCQ modelling} aims to construct a prediction model for the quality rating given the feature vectors. 
In this paper, we adopt a simple {\em linear layer} for MCQ modelling. Thus the main focus of our method is on MCQ representation. See Figure~\ref{fig:model} for an overview of these two subtasks. 

\section{Methods}\label{sec:methods}
\begin{figure*}
    \centering
    \includegraphics[width=0.7\textwidth]{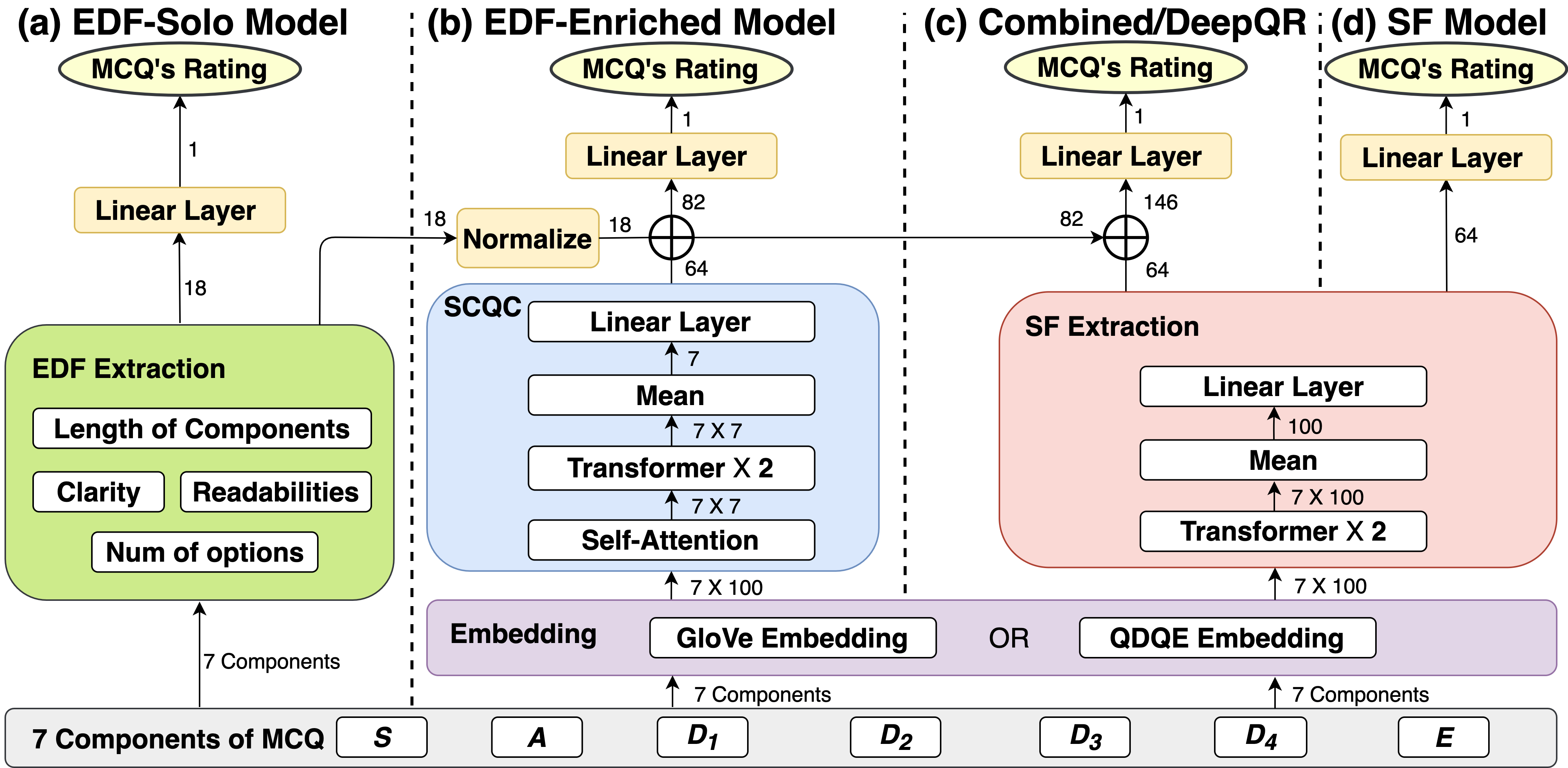}
    \caption{\footnotesize The overall architecture of the introduced AQQR models. }\label{fig:overall}
\end{figure*}

In this section, we first describe how EDF are extracted and used in AQQR. This is then followed by a description of our transformer-based SF extraction method. Last, we present our $\DeepQ$ model which combines modules developed for both preceding parts. Overall this section will present 5 AQQR models.  Fig.~\ref{fig:overall} presents an architectural overview. 

\subsection{EDF-based Models} Earlier pedagogical research have identified EDF which determine the quality of MCQ \cite{papinczak2012using,bates2014assessing,galloway2015doing}.
We introduce two EDF-based AQQR models: $\EDFsolo$ and $\EDFenriched$. The former trains linear weights of 18 EDF computed directly from the input texts. This model echoes earlier methods, e.g., winning bids of  {\sc NeurIPS 2020 Education Challenge}. The latter enriches the input of $\EDFsolo$ using {\em semantic correlations between MCQ components} (SCQC), extracted by a self-attention mechanism. 

\smallskip

\noindent {\bf Explicity-defined features.}  Details of the selected EDF are in Table~\ref{tab:edf}. The grammatical error  $\gamma_i$ is obtained using LanguageTool \cite{naber2003rule}. The nine readability indices \cite{dubay2004principles} characterise suitable reader groups of a text (e.g., by revealing the cognitive complexity required to make sense of the text) in different ways. A common feature of these indices is the use of key parameters such as average word count per sentence and number of syllables per word.  

\smallskip

\noindent {\bf The $\EDFsolo$ model.} To measure the quality of a question, classical models generally use manually-defined linear transformations on a set of EDF. In the $\EDFsolo$ model, we also use a linear transformation (of the 18 selected EDF) but the weights for the features are trained using linear regression with MSE (mean square error) loss. See Fig.~\ref{fig:overall}(a): For MCQ $M_i$, the predicted quality rating is
\begin{eqnarray}\label{eqn:linear}
  \begin{aligned}
   \resizebox{0.18\textwidth}{!}{$\hat{r}_i = 
    \begin{array}{ccc}
        \overrightarrow{w_\linear} \cdot
      	\overrightarrow{\mathrm{EDF}}_i
       + b_\linear
    \end{array}$}
  \end{aligned}
\end{eqnarray} where $\overrightarrow{\mathrm{EDF}} =  [n_{\mathrm{op}}, \nu_S, \ldots, \nu_E, \gamma, \rd_1, \ldots, \rd_9]$, $\overrightarrow{w_\linear}$ \& $b_\linear$ are (18-dim) trainable weights \& bias, resp. 


\begin{table}[h]\centering
\caption{\footnotesize Explicitly defined features.}\label{tab:edf} \footnotesize
\scalebox{0.8}{\begin{tabularx}{0.46\textwidth}{l|X}
        \toprule
        {\bf Feature Type}\hspace*{-1mm} & {\bf Features} \\
        \midrule
        {\em Options} & Number of options $n_{\mathrm{op},i}$    \\
        \hline
        {\em Length} & word counts $\nu_C$, where $C$ is taken from $\{S_i,A_i,D_{i,1},\ldots,D_{i,4},E_i\}$   \\
        \hline
        {\em Correctness} & Grammatical error  rate $\gamma_i$\\
        \hline
        {\em Readability} & Flesch reading ease\hspace*{-1mm} $\rd_{i,1}$; Flesch–Kincaid $\rd_{i,2}$; fog $\rd_{i,3}$; Coleman–Liau $\rd_{i,4}$; Linsear write formula $\rd_{i,5}$; Automated readability index $\rd_{i,6}$; Spache $\rd_{i,7}$; Dale–Chall $\rd_{i,8}$; SMOG $\rd_{i,9}$ \cite{dubay2004principles}\hspace*{-1mm}\\
        \bottomrule
    \end{tabularx}}
\end{table}

\noindent {\bf Semantic correlation of MCQ components.} 
The quality of distractors of an MCQ should be assessed in the context of other components. Indeed, a ``good'' distractor is expected to bear certain syntactic or semantic correlations with the question stem, correct answer, and possibly other distractors. We thus design SCQC to capture these correlations; See  Figure~\ref{fig:overall} (b). The input to SCQC consists of semantic embeddings of all component: $\overrightarrow{\re}=\left[\re_S, \re_A,\ldots, \re_E\right]$, where each $\re_C\in \R^{d_\embed}$ is a $d_\embed$-dim embedding of component $C\in \{S,\ldots,E\}$. These embeddings are assumed to be produced by a separate algorithm (See below).
SCQC utilises a self-attention mechanism to interpret correlations: 
\begin{eqnarray}\label{eqn:attention}
 \begin{aligned}
   \mathrm{Co} &= \Attention\left(\overrightarrow{\re}, \overrightarrow{\re}\right),\\
\Attention(V, Q) &= \softmax\left(V^\transpose WQ\right),
 \end{aligned}
\end{eqnarray}
where $Q$ is a {\em query sequence} on the {\em context sequence} $V$, $W\in \R^{d_\embed\times d_\embed}$ trainable weight matrix, and the attention score matrix $\mathrm{Co}\in \R^{7\times 7}$ represents component-wise correlations. We then encode this matrix by a 2-layer transformer encoder. Each transformer layer contains two sub-layers: a self-attention mechanism and a feed-forward layer, each of which has a residual connection before normalisation  \cite{vaswani2017attention}. Thus the output of each sub-layer is $\mathrm{LayerNorm}(X + \mathrm{Sublayer}(X))$ where $X$ is input,  $\mathrm{LayerNorm}$ is the normalisation function, and $\mathrm{Sublayer}$ is either self-attention or feed-forward as described in \eqref{eqn:selfattention} and \eqref{eqn:ffn}, respectively. Set the query, key, and value matrices as $Q=XW^Q, K=XW^K, V=XW^V$ where $W^Q, W^K, W^V$ are $d_\trans\times d_\trans$ trainable matrices, respectively. 
\begin{eqnarray}\label{eqn:selfattention}
 \begin{aligned}
 \resizebox{0.4\textwidth}{!}{$ \selfatt(Q,K,V)=\softmax\left(\frac{QK^\transpose}{\sqrt{d_\trans}}\right)V$} 
 \end{aligned}
\end{eqnarray}
\begin{eqnarray}\label{eqn:ffn}
 \begin{aligned}
\resizebox{0.42\textwidth}{!}{$\FeedForward(X)=\max\left(0, XW_1 + \vec{b_1}\right)W_2 + \vec{b_2}$}
 \end{aligned}
\end{eqnarray}
where $W_1,W_2$ are trainable weights and $\vec{b_1}$ and $\vec{b_2}$ are biases. 
The input to the first transformer layer is the correlation matrix $\mathrm{Co}$ and the next layer's input is the output of the previous layer. 
The output of the 2-layer transformer is a $7\times 7$ matrix representing the {\em encoded} correlations.
Finally, we compute the average attention score for each component which is fed into a linear layer to produce the SCQC output. Note that the parameters of SCQC are trained using prediction loss and thus SCQC captures in some sense the impact of each component to the question quality. We will showcase the ability of SCQC using a case study in Section~\ref{sec:analysis}.

\smallskip

\noindent{\bf The $\EDFenriched$ model.} See Figure~\ref{fig:overall}(b). The model concatenates the normalized EDF  with SCQC output before applying a linear layer (similar to \eqref{eqn:linear}) to obtain the rating prediction. 
We fix the popular GloVe word embeddings \cite{pennington-etal-2014-glove} to represent the MCQ components as inputs to the SCQC module. GloVe was trained with local as well as global statistics of a corpus and is able to capture semantic similarity using much lower dimensional vectors than other popular word embeddings. 

\subsection{SF-based Model}
Given the success of neural networks in building rich representations in multiple NLP tasks, it is reasonable to expect that deep semantic information captured by such models may also serve the purpose of AQQR. This section presents our method to extract such semantic information. Just as for SCQC, we use a 2-layer transformer encoder. The input of the encoder is word embeddings of the MCQ components. The transformer consists of two multi-head self-attention layers \cite{vaswani2017attention}:
%
\begin{eqnarray}\label{eqn:multihead} \footnotesize
\begin{aligned}
    \MultiHead(Q,K,V) = \left(\bigoplus_{j=1}^h \head_j\right) W^O \\
   \head_j = \selfatt\left(QW^Q_j,KW^K_j,VW^V_j\right).
 \end{aligned}
\end{eqnarray}
Here, $\Concat$ is concatenation, 
$W^Q_j, W^K_j, W^V_j \in \R^{d_{\trans}\times (d_{\trans}/h)}$, $W^O \in \mathbb{R}^{d_{\trans}\times{d_{\trans}}}$, and $h$ the head count (set $h=4$). 
Having multiple heads allows the discovery of richer information as different heads may focus on different aspects of the data. We then average the obtained representation and apply a linear transformation to get the final SF representation.

Figure~\ref{fig:overall}(d) summarises the architecture of the $\SFglove$ model. We again adopt GloVe as input embedding to the SF extraction module. After obtaining the SF representation, the $\SFglove$ model applies a final linear layer (similarly to \eqref{eqn:linear}) to obtain the predicted quality rating. We mention that efficiency amounts to a key advantage of our method. Indeed, it is straightforward to train (heavyweight) models such as $\RoBERTa$ and $\SBERT$ from the input corpus. Yet, we will demonstrate in Sec.\ref{sec:result} that this does not improve performance while incurring  heavier training costs.  

\subsection{Models Combining EDF and SF} 
This section presents two models that combine the EDF and SF in the hope to make the best use of the extracted features. 

\paragraph*{\bf The $\Combined$ model.} Following Figure~\ref{fig:overall}(c), the model takes the normalised EDF concatenated with SCQC, 
which is then concatenated with the extracted SF above. The combined vector is applied a linear transformation as in the models above to produce a quality rating prediction. The model parameters of SCQC and the SF extraction module are trained using the prediction loss. The input embedding to both SCQC and the SF extraction modules are GloVe, which is obtained without quality consideration. 
This inspires us to  fine-tune GloVe to strengthen the representations of MCQ.

\paragraph*{\bf Quality-driven question embedding and $\DeepQ$ model.} The $\DeepQ$ model differs from $\Combined$ in that it adopts QDQE for its input instead of GloVe embeddings. QDQE builds question representations while taking into account their quality rating. For this, we adopt a {\em (supervised) contrastive learning} algorithm to fine-tune a baseline language model. Contrastive learning gained considerable interest recently 
as a generic representation learning framework \cite{chen2020simple,giorgi2020declutr}. Supervised contrastive learning leverages label information in a dataset \cite{GunelDCS21}. In a nutshell, QDQE fine-tunes question embeddings derived from GloVe with a one-layer transformer encoder, so as to push apart representations of MCQ with ``unmatched quality ratings'', while pulling together representations of MCQ with ``matching quality ratings''. 

More specifically, we build 
a dataset $\mathcal{D}_{\mathrm{CL}}$ consisting of triples of questions.
Assume our training set contains MCQ $M_1,\ldots, M_n$, sorted in ascending order of quality rating. Set $S_L=\{M_1,\ldots,M_c\}$ and $S_H=\{M_{n-c+1},\ldots,M_n\}$ where $c<n/2$ is a fixed integer. The dataset set $\mathcal{D}_{\mathrm{CL}}$ contains triples of the form $(a,pos,neg)$: For each $(a,pos)\in S^2_H$ ($S^2_L$), choose a random $n_{a,pos}\in S_L$ ($S_H$). Then $\mathcal{D}_{\mathrm{CL}}=\{(a,pos,neg)\mid (a,pos)\in S^2_L\cup S^2_H, neg=n_{a,pos}\}$. Note that there are in total $2c(c-1)$ triples.

\begin{figure}
    \centering
    \includegraphics[width=0.44\textwidth]{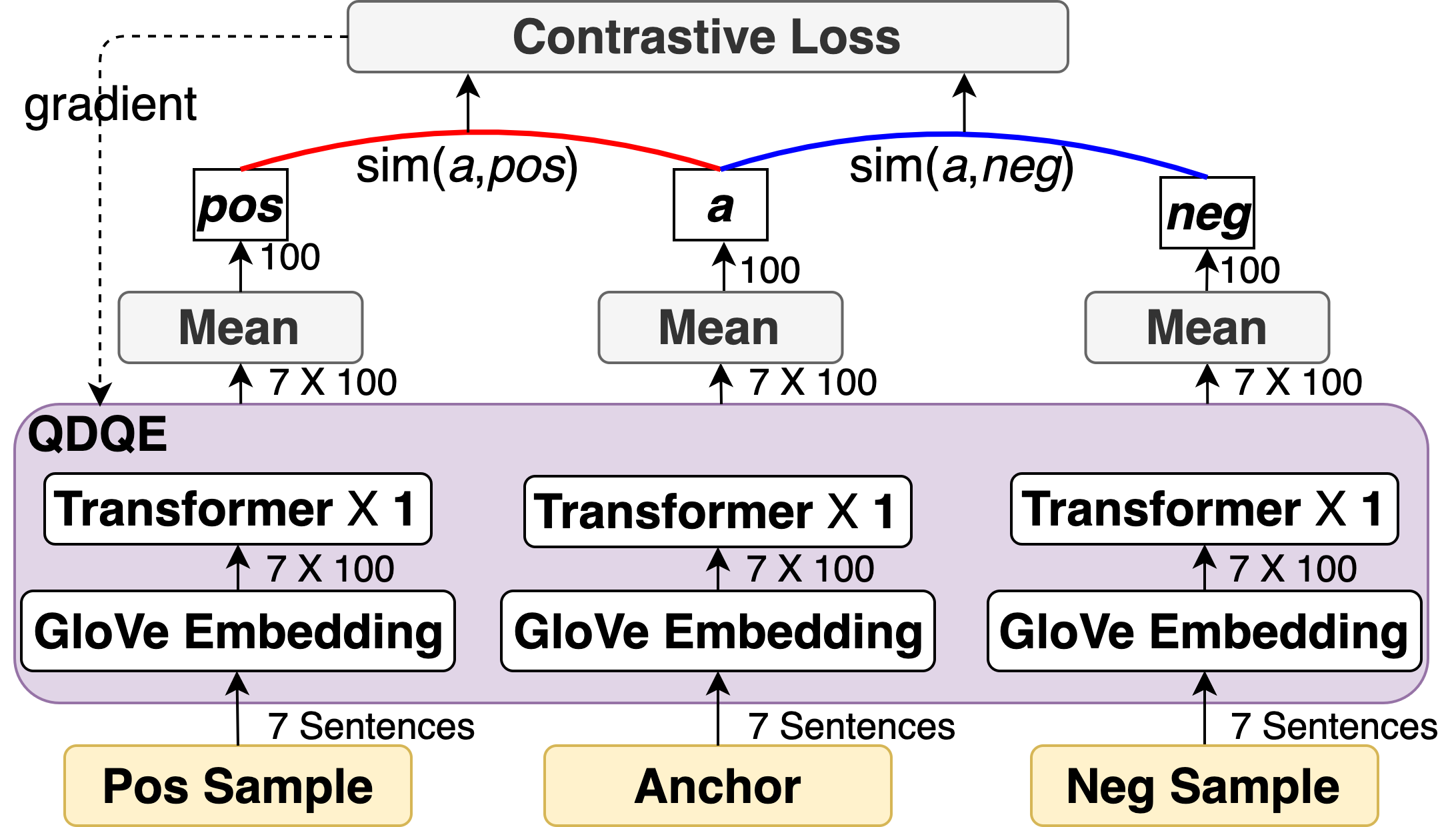}
    \caption{\footnotesize The QDQE algorithm builds MCQ representations using (supervised) contrastive learning.}\label{fig:qdqe}
\end{figure}
%
%
Figure~\ref{fig:qdqe} shows the model for training QDQE. 
The model starts from $d$-dim GloVe embeddings of the MCQ components for questions in a $(a, pos, neg)$ triple. These embeddings are encoded by a transformer that produces for each question a $7\times d$ encoded matrix. A $d$-dim vector is then computed by taking the mean of each column of the encoded matrix for each question in the triple. The contrastive loss is the InfoNCE function $\mathcal{L}_{\mathrm{IN}}$ \cite{oord2018representation}:
\begin{equation}\label{eqn:infonce} 
\resizebox{0.4\textwidth}{!}{$\mathcal{L}_{\mathrm{IN}} = -\log \frac{\exp(sim(a, pos) / \tau)}{\exp(sim(a, pos) / \tau) +  \exp(sim(a, neg) / \tau)}$}
\end{equation}
where $sim$ used here is cosine similarity, and $\tau=0.07$ following MoCo \cite{he2019momentum}. In this way, we hope the distance between the QDQE vectors $\vec a$ and $\vec p$ becomes less than the distances between  $\vec a$ and $\vec n$, as well as between $\vec p$ and $\vec n$. 

\section{Experiments}\label{sec:experiment}

\subsection{Experiment Setup}

\noindent {\bf Datasets.} Our eight PeerWise datasets are taken from a law course and seven medicine courses (M1,4,7 are the same course of a university in different school years, similar for M2,3,5,6). Each dataset contains MCQ that are presented as in Section~\ref{sec:problem} where the average rating is the ground truth. To ensure the reliability, only questions receive at least 10 ratings are included. See Table~\ref{tab:data} for the final datasets details. 

\begin{table}\footnotesize \centering
\caption{\footnotesize The details of the eight PeerWise datasets.}\label{tab:data}
\scalebox{0.84}{\begin{tabular}{c|cccc}
\toprule
Subject & Law & M1 & M2 & M3 \\ \midrule
\# MCQ & 3,834 & 1,747 & 1,509 & 2,021 \\ 
\# ratings & 72,753 & 141,889 & 92,607 & 152,387 \\ 
Ratings per MCQ & 18.97 & 81.21 & 61.36 & 75.40 \\ 
Av. stem length & 101.75 & 198.29 & 112.21 & 130.93 \\ 
\bottomrule \toprule
 Subject & M4 & M5 & M6 & M7 \\ \midrule
 \# MCQ & 1,205 & 2,879 & 1,250 &905 \\ 
\# ratings & 143,654 & 219,084 & 91,719 &109,549 \\ 
Ratings per MCQ & 119.21 & 76.09 & 73.37 &121.04 \\ 
Av. stem length & 246.96 & 163.40 & 192.45 & 190.25 \\ \bottomrule
\end{tabular}}
\end{table}


\smallskip

\noindent {\bf Models.} We test five AQQR models ($\EDFsolo$, $\EDFenriched$, $\SFglove$, $\Combined$, $\DeepQ$) and two benchmark models $\RoBERTa$ and $\SBERT$. The benchmark models are implemented using roberta-base and sentence-transformers/paraphrase-distilroberta-base-v1 from {\sc Huggingface} \cite{wolf2019huggingface}, resp. and are fine-tuned on AQQR. For these benchmarks we combine all MCQ components as a single input. We split each dataset into training, validation, test set by 8:1:1. We use a seed of 2021, and train our models using Adam optimizer \cite{DBLP:journals/corr/KingmaB14} for 50 epochs from which the epoch achieves the lowest validation loss (MSE: {\em mean square error}) is chosen.

\smallskip

\noindent {\bf Performance measures.} We use both MSE and ACC to measure AQQR performance. 
For ACC, we count a predicted rating as {\em correct} when $|r_i-\hat{r}_i|\leq 0.25$ (recall $r_{i}$ \& $\hat{r}_{i}$ are resp. the ground truth \& predicted labels) and define ACC as the fraction of correct predictions in the test set. 
While ACC offers insight on the model's ability, MSE can be seen as a more reliable metric. 

\smallskip

\noindent {\bf Hyper-parameters.} We set the batch size to 16 and the initial learning rate to $1e-3$. For the optimizer learning rate scheduler,  we set step size to 3 and gamma to 0.7 for the optimizer learning rate scheduler.  
We set the dropout to 0.5 for our AQQR models and to 0.1 for the two benchmarks. For QDQE, we train a model separately for each course dataset with c = 80 and 20 for the train and validation dataset respectively. The hyper-parameter settings are inherited from benchmark models except the batch size is 1.


\smallskip

\noindent {\bf Experiments design.} We conduct experiments in three stages to verify modules within our design. 
We first compare $\EDFsolo$ with $\EDFenriched$ to highlight the power of $\SCQC$, and then compare $\SFglove$ with  $\RoBERTa$ and $\SBERT$ to showcase our SF extraction module. We last demonstrate the value of the combined models $\Combined$ and $\DeepQ$ hoping to validate the use of QDQE.

\subsection{Results}\label{sec:result} All experiments are conducted on NVIDIA 460.84 Linux Driver. The CUDA version is 11.2 and the Quadro RTX 8000 with 48 GB GPU memory for our experiment. The CPU version is Intel(R) Xeon(R) Gold 5218 CPU @ 2.30GHz and 16 cores. 
The results are shown in Table~\ref{tab:experiment}. We make the following observations: (1) As seen from the first two rows, $\EDFenriched$ made a substantial improvement from $\EDFsolo$ across all datasets, achieving almost $80\%$ less MSE for M7 and $43\%$ less MSE for M6 less. This reflects the result after enriching the EDF with SCQC. (2) Among the SF-based models, $\SFglove$ outperforms the benchmarks of $\RoBERTa$ and $\SBERT$ in most of the cases, and performs comparably well as $\EDFenriched$. This demonstrates that deep learning captures sufficient semantic information to express question quality. (3) In general, the models that combine EDF and SF achieve the best accuracy. This is despite the fact that $\Combined$'s lead is not conclusive on some of the datasets, e.g., getting higher loss than other models in Law and M1. This is somewhat surprising as enlarging the input features does not necessarily boost the performance. Nevertheless, $\DeepQ$ scores the best performance across all datasets in terms of both MSE and ACC. The only two exceptions are the ACC scores on Law and M5 which are within $1.4\%$ and $0.4\%$ of the best scores respectively. In most of the datasets, $\DeepQ$ achieves $2\%$+ better than the next best model in terms of ACC. This demonstrates the benefit of using QDQE as the input sentence embedding.

\begin{table*}
\centering
\caption{\footnotesize AQQR performance (MSE \& ACC(\%)) of seven models on eight PeerWise datasets.}
\label{tab:experiment}
\scalebox{0.75}{
\begin{tabular}{@{}l|cc|cc|cc|cc|cc|cc|cc|cc@{}}
\toprule
Dataset& \multicolumn{2}{c}{Law} & \multicolumn{2}{c}{M1} & \multicolumn{2}{c}{M2} & \multicolumn{2}{c}{M3} & \multicolumn{2}{c}{M4} & \multicolumn{2}{c}{M5} & \multicolumn{2}{c}{M6} & \multicolumn{2}{c}{M7} \\ \midrule
Model & MSE & ACC & MSE & ACC & MSE & ACC & MSE & ACC & MSE & ACC & MSE & ACC & MSE & ACC & MSE & ACC \\ \hline
$\EDFsolo$ & 0.119 & 53.64 & 0.104 & 61.14 & 0.133 & 61.58 & 0.062 & 71.42 & 0.064 & 67.76 & 0.037 & 84.37  & 0.084 & 69.60 & 0.206 & 75.82 \\
$\mathsf{EDF\mhyphen Enr.}$ & {0.115} & {53.91} & {0.079} & {78.29} & {0.097} & {67.55} & {0.041} & {83.74} & {0.041} & {82.64} & {0.030} & \textbf{90.62} & {0.048} & {76.00} & {0.042} & {84.62} \\ \hline
$\SFglove$ & \textbf{0.107} & \textbf{57.55} & {0.064} & 77.71 & {0.103} & {64.90} & {0.038} &  86.21 & {0.038} &  {84.29} & {0.030} &  \textbf{90.62} & {0.044} &  77.60 & {0.038} &  85.71 \\
$\RoBERTa$ & 0.117 & 54.68 & 0.064 & {78.85} & 0.109 & 62.91 & 0.042 & {86.69} & 0.040 & 83.47 & 0.032 & 89.23 & 0.049 & {80.80} & 0.042 & 86.81 \\
$\SBERT$ & 0.117 & 54.68 & 0.064 & {78.85} & 0.113 & 56.95 & 0.043 & 80.29 & 0.040 & 83.47 & 0.033 & 89.23 & 0.050 & {80.80} & 0.040 & {87.91} \\
\hline
$\Combined$ & 0.122 & 52.08 & 0.071 & 77.14 & 0.097 & 66.89 & 0.038 & 87.19 & 0.040 & 84.30 & 0.030 & \textbf{90.62} & \textbf{0.039} & 82.40 & \textbf{0.034} & 87.91 \\
$\DeepQ$ & \textbf{0.107} & {56.77} & \textbf{0.060} & \textbf{80.57} & \textbf{0.093} & \textbf{68.87} & \textbf{0.036} & \textbf{88.18} & \textbf{0.037} & \textbf{85.95} & \textbf{0.029} & 90.27 & \textbf{0.039} & \textbf{84.80} & \textbf{0.034} & \textbf{90.11} \\ \bottomrule
\end{tabular}}
\end{table*}

To illustrate the computational efficiency of our transformer-based SF extraction module, Table~\ref{tab:time} shows the average training time per epoch of the $\RoBERTa$, $\SBERT$ and $\DeepQ$ model. In all but one dataset, $\DeepQ$ outperforms the benchmarks. 
\begin{table}
\centering
\caption{\footnotesize Average training time per epoch of $\RoBERTa$, $\SBERT$ and $\DeepQ$ (in second).}\label{tab:time}
\scalebox{0.8}{
\begin{tabular}{@{}lcccccccc@{}}
\toprule
Model & Law & M1 & M2 & M3 & M4 & M5 & M6 & M7 \\ \midrule
$\RoBERTa$ & 109.66 \hspace*{-2mm}& 50.00 \hspace*{-2mm} & 43.33 \hspace*{-2mm} & 57.00 \hspace*{-2mm} & 34.00 \hspace*{-2mm} & 86.00 \hspace*{-2mm} & 35.00 \hspace*{-2mm} & 26.00 \hspace*{-2mm} \\
$\SBERT$ & 55.00 \hspace*{-2mm}& \textbf{25.00} \hspace*{-2mm}& 22.66 \hspace*{-2mm} & 29.00 \hspace*{-2mm} & 18.00 \hspace*{-2mm} & 44.00 \hspace*{-2mm} & 18.00 \hspace*{-2mm} & 14.00 \hspace*{-2mm} \\
$\DeepQ$ & \textbf{44.33} \hspace*{-2mm}& 27.66 \hspace*{-2mm} & \textbf{17.66} \hspace*{-2mm}& \textbf{28.00}\hspace*{-2mm} & \textbf{16.33} \hspace*{-2mm} & \textbf{40.00} \hspace*{-2mm} & \textbf{17.33} \hspace*{-2mm}& \textbf{12.66} \hspace*{-2mm}\\ \bottomrule
\end{tabular}}
\end{table}

\subsection{Analysis and Discussion}\label{sec:analysis}
We present further analysis on our $\DeepQ$ model. 

\begin{figure}
\centering
\includegraphics[width=0.37\textwidth]{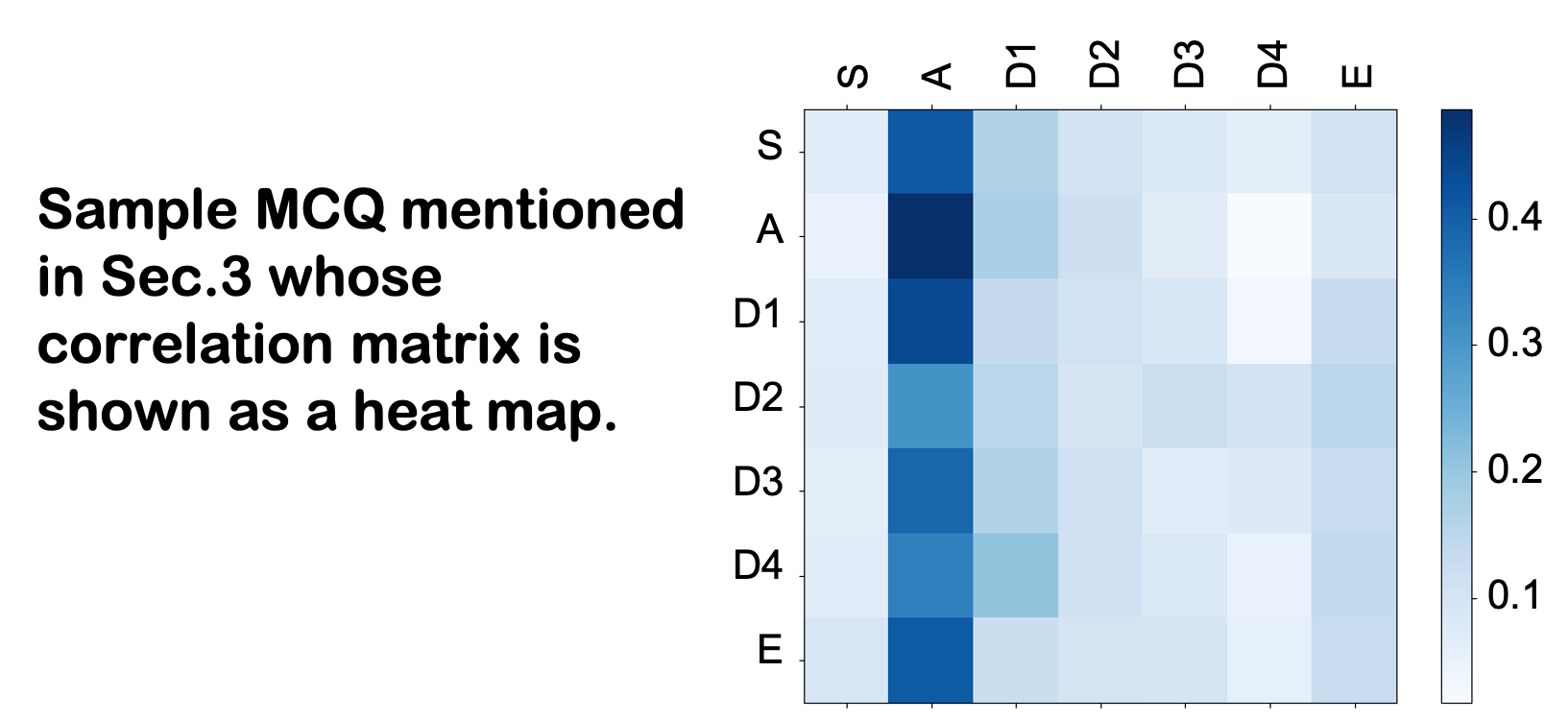}
\caption{\footnotesize Visualisation of SCQC attention matrix.}
\label{fig:heatmap}
\end{figure}

\paragraph*{\bf Case study on SCQC.} One benefit of our SCQC module is its explanatory ability. By visualising the self-attention matrix, we are able to observe correlations among MCQ components which are calibrated to reveal quality. Fig.~\ref{fig:heatmap} displays the case study whose correlation matrix is shown as a heat map. Darker blue cells indicate a higher correlation between the components. The diagram shows high correlations between the stem, and the distractor 1 ``Hallucinations'' with the answer ``Schizophrenia'', which help to reveal question quality. This diagram hints that SCQC facilitates a question to capture meaningful insights for the model explanation. 

\smallskip

\begin{figure}[ht]
\resizebox{0.46\textwidth}{!}{\includegraphics{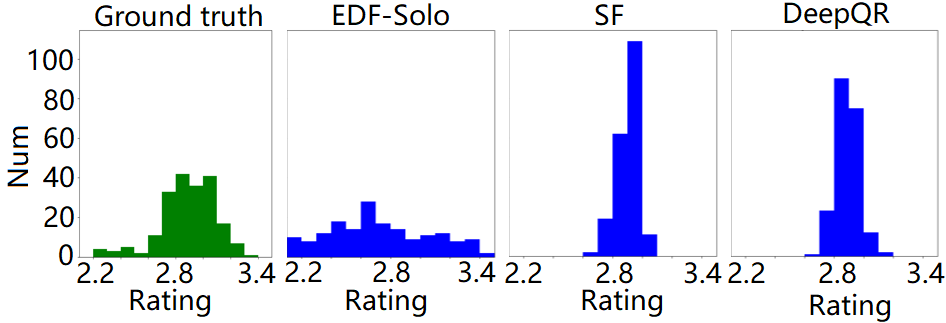}}
\caption{\footnotesize Ground truth and prediction distributions on M3.}\label{fig:predict-rating-ground-truth}
\end{figure}

\noindent {\bf Comparison of rating distributions.} Fig.~\ref{fig:predict-rating-ground-truth} compares the four rating distributions on M3 dataset as a case study: ground truth, $\EDFsolo$, $\SFglove$, and $\DeepQ$ predictions. The histogram displays the number of questions whose ratings fall within different intervals. While ratings produced by $\EDFsolo$ are too evenly distributed, and those by $\SFglove$ concentrate too much in one rating interval, the rating distribution obtained by $\DeepQ$ strikes a balance between the two and most resembles the ground truth. 


\smallskip

\noindent {\bf Identifying questions with high- (low-)quality.} A model that can detect questions with exceptionally low or high quality could be used as either a question filter (to eliminate low-quality questions) or a question recommender (to promote high-quality questions). We verify these abilities for $\DeepQ$: Call a question ``high-quality'' (or ``low-quality'') if its rating falls one standard deviation above (or below) the mean of its dataset. Thus, these questions account for roughly $15\%$ of total samples assuming the ratings are normally distributed. Suppose we classify a test sample by the same rule above, but according to the predicted rating. We measure {\em classification accuracy}, namely, the proportion of questions in the test set that are correctly classified as ``high'' or ``not high''(and ``low'' or ``not low''). See Table~\ref{tab:ACC-SD}: the results show that $\DeepQ$ achieves reasonably high classification accuracy for both types of questions over all datasets. 

\begin{table}
\centering
\caption{\footnotesize Classification accuracy (\%) for high- and low-quality questions by $\DeepQ$.}
\label{tab:ACC-SD}
\scalebox{0.75}{
\begin{tabular}{@{}lcccccccc@{}}
\toprule
Cl. ACC & Law & M1 & M2 & M3 & M4 & M5 & M6 & M7 \\ \midrule
Low & 76.30 & 83.42 & 80.79 & 77.83 & 80.99 & 85.06 & 84.80 & 83.51 \\
High & 75.78 & 82.85 & 87.41 & 84.23 & 88.42 & 80.55 & 80.80 & 86.81 \\ \bottomrule
\end{tabular}}
\end{table}

\paragraph*{Error analysis.} 
Although $\DeepQ$ achieves superior performance than other models, it nevertheless predicts falsely on many MCQs. In particular, the performance on the Law dataset is considerably worse than on other datasets. Many factors potentially contribute to this (see Table~\ref{tab:data}): (1) Law has the lowest number of ratings per MCQ (18.97) which casts a doubt on its reliability. (2) Question stems in Law have the shortest average character-level length (101.75) which could affect performance. 
(3) A larger proportion of MCQs in Law are numerical (e.g. on taxation). 

\begin{figure}[ht]
\resizebox{!}{3.15cm}{\includegraphics[width=4.6cm]{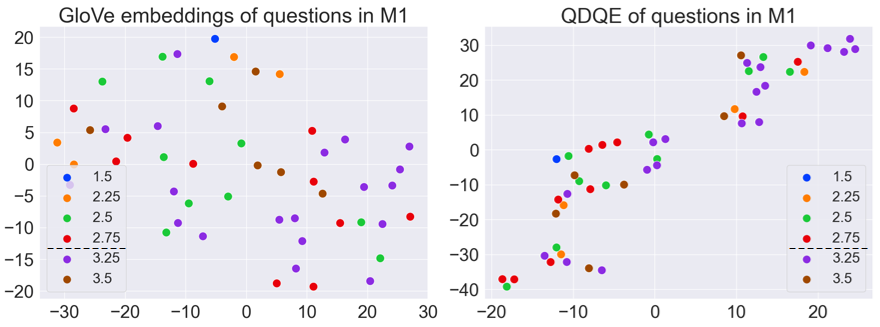}}
\caption{\footnotesize t-SNE \cite{van2008visualizing} embedding visualization of GloVe (left) and QDQE (right) where points are questions and colours indicate rating categories.}\label{fig:QDQEvisualization}
\end{figure}

Another potential source of inaccuracies to our model lies in the MCQ representations.
Fig.~\ref{fig:QDQEvisualization} visualises the GloVe and QDQE question embeddings of 50 questions from M1. While it is apparent that QDQE results in a strengthened clustering effect (e.g. highly rated purple points tend to cluster in the upper right quadrant while poorly-rated red and green points cluster on the left), the resulting clustering is not entirely determined by rating categories. This shows that QDQE alone is not sufficient to capture quality rating, which is not surprising as this way of obtaining question embeddings does not account for contents in the entire course. 


\section{Conclusions and Future Work}

This paper investigates AQQR using tools from deep learning. We propose the $\DeepQ$ model that combines EDF and SF sources extracted by transformer networks, as well as contrastive learning-based question embeddings. Empirical results using eight PeerWise datasets validate the superior performance of $\DeepQ$ over six comparative models. Future work includes improving the model's accuracy through better question embedding schemes
and incorporating domain-specific knowledge. In addition, when training our model we aim to improve the quality of the input data (aggregated student ratings) using effective consensus approaches \cite{abdi2021evaluating,darvishi2021employing}. 
Finally, we aim to evaluate the use of $\DeepQ$ in practice for recommending high-quality questions to students who are engaged in practice tasks.


\bibliography{main} 

\end{document}